%% file: neurips_2025.tex
\documentclass{article}

\usepackage[preprint]{neurips_2025}

\usepackage[utf8]{inputenc} % allow utf-8 input
\usepackage[T1]{fontenc}    % use 8-bit T1 fonts
\usepackage{hyperref}       % hyperlinks
\usepackage{url}            % simple URL typesetting
\usepackage{booktabs}       % professional-quality tables
\usepackage{amsfonts}       % blackboard math symbols
\usepackage{nicefrac}       % compact symbols for 1/2, etc.
\usepackage{microtype}      % microtypography
\usepackage{xcolor}         % colors

\usepackage{booktabs}       % professional-quality tables
\usepackage{multirow}       % for multi-row table cells
\usepackage{wrapfig}
\usepackage{enumitem}
\usepackage{amsmath}
\usepackage{amsfonts}       % blackboard math symbols
\usepackage{algpseudocode}  % algorithmic environment
\usepackage{algorithm}
\usepackage{graphicx} 

\usepackage{xcolor} % 用于颜色支持
\definecolor{academicdarkred}{rgb}{0.6, 0.1, 0.1}

\usepackage{amsmath}  % For align* and advanced math
\usepackage{amssymb}  % For symbols like \mathbb
\usepackage{adjustbox}
\usepackage{tcolorbox}
\usepackage{arydshln}

\usepackage{pifont} % 提供对号和叉号
\usepackage{xcolor} % 提供颜色支持

\usepackage{subcaption}
\usepackage{float}
\usepackage{caption}

\newcommand{\xmark}{\textcolor{red}{\ding{55}}}   % 红色叉号

\definecolor{myblue}{HTML}{4E84C4}
\definecolor{myred}{HTML}{B02418}
\definecolor{mygreen}{HTML}{34692E}
\definecolor{myorange}{HTML}{DA7842}
\definecolor{paperblue}{HTML}{077dea}
\definecolor{babyblue}{HTML}{E3EDF7} 

\definecolor{mintleaf}{RGB}{0, 184, 148}
\definecolor{dm-blue-500}{RGB}{0, 69, 177}
\definecolor{dm-purple-500}{RGB}{105,50,230}
\definecolor{mysilver}{RGB}{128,129,128}
\definecolor{my_green}{RGB}{0, 176, 80}
\definecolor{my_yellow}{RGB}{255,165,0}
\definecolor{my_red}{RGB}{255, 0, 0}
\definecolor{my_purple}{RGB}{126, 100, 158}
\definecolor{my_blue}{RGB}{49, 133, 155}
\definecolor{case_purple}{RGB}{160, 43, 147}
\definecolor{case_blue}{RGB}{15, 158, 213}

\usepackage{wrapfig,booktabs}
\title{SPA-RL: Reinforcing LLM Agents via Stepwise Progress Attribution}

\author{
Hanlin Wang, ~~Chak Tou Leong, ~~Jiashuo Wang, ~~Jian Wang\thanks{Corresponding author.}, ~~Wenjie Li \\
Department of Computing, The Hong Kong Polytechnic University \\
\texttt{\{hanlin-henry.wang,chak-tou.leong\}@connect.polyu.hk} \\
\texttt{jian51.wang@polyu.edu.hk} \\
\texttt{\{csjwang,cswjli\}@comp.polyu.edu.hk}
}

\begin{document}

\maketitle

\begin{abstract}

\input{sections/0_abstract}

\end{abstract}

\input{sections/1_introduction}
\input{sections/2_prelim}
\input{sections/3_method}

\input{sections/4_experiment}
\input{sections/5_analyses}
\input{sections/6_related_work}
\input{sections/7_conclusion}

% \newpage
\bibliography{neurips_2025}
\bibliographystyle{unsrt}

% \bibliographystyle{unsrt}
% \bibliography{reference.bib}

%%%%%%%%%%%%%%%%%%%%%%%%%%%%%%%%%%%%%%%%%%%%%%%%%%%%%%%%%%%%

\newpage
\appendix
\input{sections/appendix}

\end{document}

%% file: sections/0_abstract.tex
Reinforcement learning (RL) holds significant promise for training LLM agents to handle complex, goal-oriented tasks that require multi-step interactions with external environments.
However, a critical challenge when applying RL to these agentic tasks arises from delayed rewards: \textit{feedback signals are typically available only after the entire task is completed}.
This makes it non-trivial to assign delayed rewards to earlier actions, providing insufficient guidance regarding environmental constraints and hindering agent training.
In this work, we draw on the insight that the ultimate completion of a task emerges from the cumulative progress an agent makes across individual steps.
We propose \textbf{S}tepwise \textbf{P}rogress \textbf{A}ttribution (\textbf{SPA}), a general reward redistribution framework that decomposes the final reward into stepwise contributions, each reflecting its incremental progress toward overall task completion.
To achieve this, we train a progress estimator that accumulates stepwise contributions over a trajectory to match the task completion.
During policy optimization, we combine the estimated per-step contribution with a grounding signal for actions executed in the environment as the fine-grained, intermediate reward for effective agent training.
Extensive experiments on common agent benchmarks (including Webshop, ALFWorld, and VirtualHome) demonstrate that SPA consistently outperforms the state-of-the-art method in both success rate (+2.5\% on average) and grounding accuracy (+1.9\% on average). 
Further analyses demonstrate that our method remarkably provides more effective intermediate rewards for RL training.
Our code is available at ~\url{https://github.com/WangHanLinHenry/SPA-RL-Agent}.

%% file: sections/1_introduction.tex
\section{Introduction}

Recent advances in large language models (LLMs) have demonstrated their impressive capability to serve as intelligent agents, yielding a wide range of application domains such as web agents~\citep{zhou2024archer,zhou2025sweet}, search agents~\citep{jin2025search,wang2025otc,li2025search}, code agents~\citep{yang2023intercode,zhang2024codeagent}, and embodied agents~\citep{wang2025ragen,wang2024e2cl,wang2025steca}.
LLM-based agents often perform multi-step planning and interact with external environments to complete specific tasks.
In such goal-reaching scenarios, reinforcement learning (RL)~\citep{sutton1999policy} is widely used, as it optimizes sequential decision-making and naturally endows agents with ``explore-and-exploit'' capabilities to maximize task performance in dynamic and interactive environments.

However, a significant challenge still remains for RL training in LLM agents: agentic tasks are often long-horizon with \textit{sparse and delayed rewards}, where a reward indicating the task completion is provided only at the end of the trajectory. This nature of sparsity and delay makes it difficult to propagate the final reward signal back to the intermediate actions taken in earlier steps of the trajectory. Without such intermediate feedback, agents often struggle to distinguish which specific actions contribute positively or negatively to the final outcome.
While recent studies have increasingly adopted process supervision to address the challenges posed by delayed rewards~\citep{deng2024novice,choudhury2025process}, these approaches primarily emphasize local optimization (as summarized in Table~\ref{tab:intro}). That is, they tend to greedily optimize for actions that appear optimal in the short term, often overlooking their alignment with long-term goal achievement. 
This limitation is critical, as short-sighted decisions can ultimately lead to globally suboptimal outcomes. 
To train effective LLM agents for long-horizon tasks, it is therefore essential to assign intermediate rewards that not only guide stepwise behavior but also provide guarantees of global optimality.

% 相关工作提示要进行stepweise的信号在rl中
\begin{table}[t!]
\caption{Comparison among different RL-based training methods in the ALFWorld environment.}
\vspace{5pt}
\label{tab:intro}
\centering
\resizebox{0.98\textwidth}{!}{
\begin{tabular}{lcccc}
\toprule
\textbf{Method} &   \textbf{Primary Optimization Objective} & \textbf{Process Rewarding} &  \textbf{Success (\%)} \\
\midrule
Policy Gradient~\citep{sutton1999policy} &   Global (Maximize final reward)  & \xmark & 72.4 \\
RLOO~\citep{ahmadian2024back} &   Global (Maximize final reward) & \xmark &  71.6 \\
GRPO~\citep{guo2025deepseek} &   Global (Maximize final reward)   & \xmark &  70.9 \\
StepAgent~\citep{deng2024novice} &   Local (Optimize per-step action greedily) &  Imitation Reward & 75.4 \\
PRM4A~\citep{choudhury2025process} &   Local (Optimize per-step action greedily)  & Stepwise Value & 73.9 \\
\midrule
\textbf{SPA (Ours)} &   Global \textbf{(Maximize task completion)} & \textbf{Stepwise Progress} & \textbf{79.1} \\
\bottomrule
\end{tabular}}
\end{table}

In this work, we introduce \textbf{S}tepwise \textbf{P}rogress \textbf{A}ttribution (\textbf{SPA}), a general reward redistribution framework that effectively reinforces LLM agents by offering goal-oriented intermediate rewards.
SPA is based on the observation that completing an agentic task requires accumulated progress from individual steps, with each step contributing toward the final goal. 
This \textit{stepwise progress} serves as the basis for intermediate rewards.
To achieve this, we train a progress estimator to assign a contribution score for each step. The sum of per-step scores represents the agent's estimated task completion.
The estimator is optimized to minimize the difference between the cumulative score and the actual task completion, i.e., the observed outcome reward.
As the estimator becomes sufficiently accurate in predicting the task completion, its estimated per-step contribution becomes a valid indicator of incremental progress.
At the same time, the summation form of stepwise contributions consistently aligns the intermediate attribution objective with the ultimate task completion goal
(see Appendix~\ref{reason_resa} for theoretical analysis). 
During agent policy training, we combine the predicted stepwise progress with a \textit{grounding signal} indicating whether each step is able to be executed in the environment as the final intermediate reward, to further enhance the learning effectiveness in RL.
Extensive experiments on three common agentic benchmarks (WebShop~\citep{yao2022webshop}, ALFWorld~\citep{shridhar2020alfworld}, and VirtualHome~\citep{puig2018virtualhome}) demonstrate that our SPA consistently outperforms previous approaches. 
On average, SPA achieves a 2.5\% improvement in success rate and a 1.9\% improvement in grounding accuracy compared to the state-of-the-art method.
These results highlight the effectiveness of our method in both generating executable actions and completing tasks.
Further analyses confirm that our method yields suitable intermediate rewards more effectively for developing LLM agents.

In summary, our key contributions are as follows:
\begin{itemize}[leftmargin=*]
    \item We propose Stepwise Progress Attribution (SPA), a novel framework that attributes incremental progress to each step within a multi-step trajectory.
    By decomposing delayed rewards into stepwise contributions, SPA enables effective reward redistribution and ensures a cohesive alignment between intermediate progress and final task completion.
    \item We seamlessly integrate SPA into reinforcement learning by leveraging its predicted stepwise progress as dense intermediate rewards. Combined with grounding signals for executable actions, this approach significantly enhances the training of LLM-based agents on complex agentic tasks.
    \item Our method achieves state-of-the-art performance across several widely used agent benchmarks (including Webshop, ALFWorld, and VirtualHome), consistently outperforming existing methods in success rate and grounding accuracy. 
    Extensive analyses substantiate that SPA provides superior intermediate rewards, yielding more effective RL training outcomes.
\end{itemize}

%% file: sections/2_prelim.tex
\section{Preliminaries}

\subsection{Task Formulation}

We formalize agentic problems as a partially observable Markov decision process (POMDP) defined by the tuple $(\mathcal{U}, \mathcal{S}, \mathcal{A}, \mathcal{O}, \mathcal{T}, \mathcal{R})$, 
where $\mathcal{U}$ denotes the instruction space, $\mathcal{S}$ the state space, $\mathcal{A}$ the action space, $\mathcal{O}$ the observation space, $\mathcal{T}\colon \mathcal{S}\times \mathcal{A}\to \mathcal{S}$ is the transition function, and $\mathcal{R}\colon \mathcal{S}\times \mathcal{A}\to [0,1]$ is the reward function. Since our primary focus is on the high-level task planning capabilities of LLM agents, we restrict $\mathcal{U}$, $\mathcal{A}$, and $\mathcal{O}$ to subsets of natural‐language form.

Given a task instruction $u\in\mathcal{U}$, the LLM agent $\pi_{\theta}$ at time step $t$ takes an action $a_t\sim \pi_\theta(\cdot|u, e_{t-1})$ and receives the environmental feedback as the observation $o_t\in\mathcal{O}$. $e_{t-1}$ denotes the historical interaction trajectory $(a_1, o_1, ... , a_{t-1}, o_{t-1})$. Each action $a_t$ incurs the environment state to $s_t\in\mathcal{S}$. The interaction loop terminates when either the agent completes the task or the maximum step is reached.
The final trajectory is $e_n = (u, a_1, o_1, ..., a_n, o_n)$, where $n$ denotes the length of trajectory.

A key challenge in the above setting is that the reward function $\mathcal{R}$ typically provides a sparse signal, with meaningful rewards only at the end of a trajectory. Specifically, $r_t = 0$ for all $t < n$ and $r_n \in [0, 1]$ indicating task success or failure, respectively. This sparse structure makes it delayed and difficult to propagate final rewards to earlier steps, especially for long-horizon trajectories.

\subsection{Proximal Policy Optimization}

Proximal Policy Optimization (PPO)~\citep{schulman2017proximal} is a popular reinforcement learning algorithm widely applied in various goal-reaching tasks.
It directly maximizes the expected cumulative reward:
\begin{equation}
J(\theta) = \mathbb{E}_{\tau\sim\pi_\theta}\Bigl[\sum_{t=1}^n \gamma^{\,t-1}r_t\Bigr],
\end{equation}
where \(\pi_\theta\) is the policy with parameters \(\theta\), \(\gamma\) the discount factor, and \(r_t\) the reward at step \(t\). 

Rather than taking unconstrained gradient steps, PPO uses a clipped surrogate objective to ensure that each policy update does not deviate too far from the previous one. Specifically, it optimizes
\begin{equation}
\mathcal{L}^{\mathrm{CLIP}}(\theta) \;=\; \mathbb{E}_t\Bigl[\min\bigl(\dfrac{\pi_\theta(a_t\mid s_t)}{\pi_{\theta_{\mathrm{old}}}(a_t\mid s_t)},\hat A_t,\;\mathrm{clip}(\dfrac{\pi_\theta(a_t\mid s_t)}{\pi_{\theta_{\mathrm{old}}}(a_t\mid s_t)},1-\epsilon,1+\epsilon)\,\hat A_t\bigr)\Bigr],
\end{equation}
and the advantages \(\hat A_t\) are typically computed via Generalized Advantage Estimation (GAE)~\citep{schulman2015high}, which blends temporal-difference errors
\(\delta_t = r_t + \gamma V_\phi(s_{t+1}) - V_\phi(s_t)\) with an exponentially decaying weight \(\lambda\) to trade off bias and variance. Meanwhile, a separate value network \(V_\phi\) is trained by minimizing the mean-squared error between predicted values and empirical returns. PPO alternates between optimizing the clipped policy objective and updating the value function, yielding a robust and widely used algorithm for many goal-reaching problems, including language agentic tasks.
In this work, our LLM agent training is built upon PPO.

\subsection{Limitations of PPO in Long-Horizon Tasks with Sparse, Delayed Rewards}
\label{pre_why_no_dr}

When rewards are observed only at episode termination, i.e., \(r_t=0\) for \(t<n-1\), \(r_n\in [0,1]\), the temporal difference (TD) error \(\delta_t\) vanishes for all but the final step.  Under GAE, we have
\begin{equation}
\hat A_t
= \sum_{k=0}^{n-t-1} (\gamma\lambda)^k\,\delta_{t+k}
% = \sum_{k=0}^{n-t-1} (\gamma\lambda)^k\,\delta_{t+k} + (\gamma\lambda)^{\,n-t}\delta_n
\approx (\gamma\lambda)^{\,n-t-1}\delta_{n-1},
\label{eq:advantage}
\end{equation}
where one-step TD error, except the last two steps, could be assumed as zero in the ideal situation. The detailed analyses are presented in Appendix~\ref{more_theo_proof}. Therefore, the policy gradient is written as:
\begin{equation}
\nabla_\theta J
\approx
\mathbb{E}\Bigl[\sum_{t=1}^{n-1}\nabla_\theta\log\pi_\theta(a_t\mid s_t)\,\hat A_t\Bigr]
=
\mathbb{E}\Bigl[\sum_{t=1}^{n-1}(\gamma\lambda)^{\,n-t-1}\,\delta_{n-1}\,
\nabla_\theta\log\pi_\theta(a_t\mid s_t)\Bigr],
\end{equation}
which carries an \emph{exponentially vanishing} weight \((\gamma\lambda)^{\,n-1-t}\) on early actions whenever \(\gamma\lambda<1\). Consequently, PPO with GAE fails to propagate a sparse, delayed reward back to actions taken many steps before the end.

In sum, the combination of exponentially vanishing advantages, severe discount attenuation, and degenerate value‐function learning renders vanilla PPO ineffective for long‐horizon tasks with sparse rewards.  These foundational obstacles motivate our SPA framework, which injects intermediate supervisory signals to restore informative learning gradients throughout the trajectory.

%% file: sections/3_method.tex
\section{Method}
\label{sec:method}

\begin{figure}[t!]
  \centering
  \includegraphics[width=\textwidth]{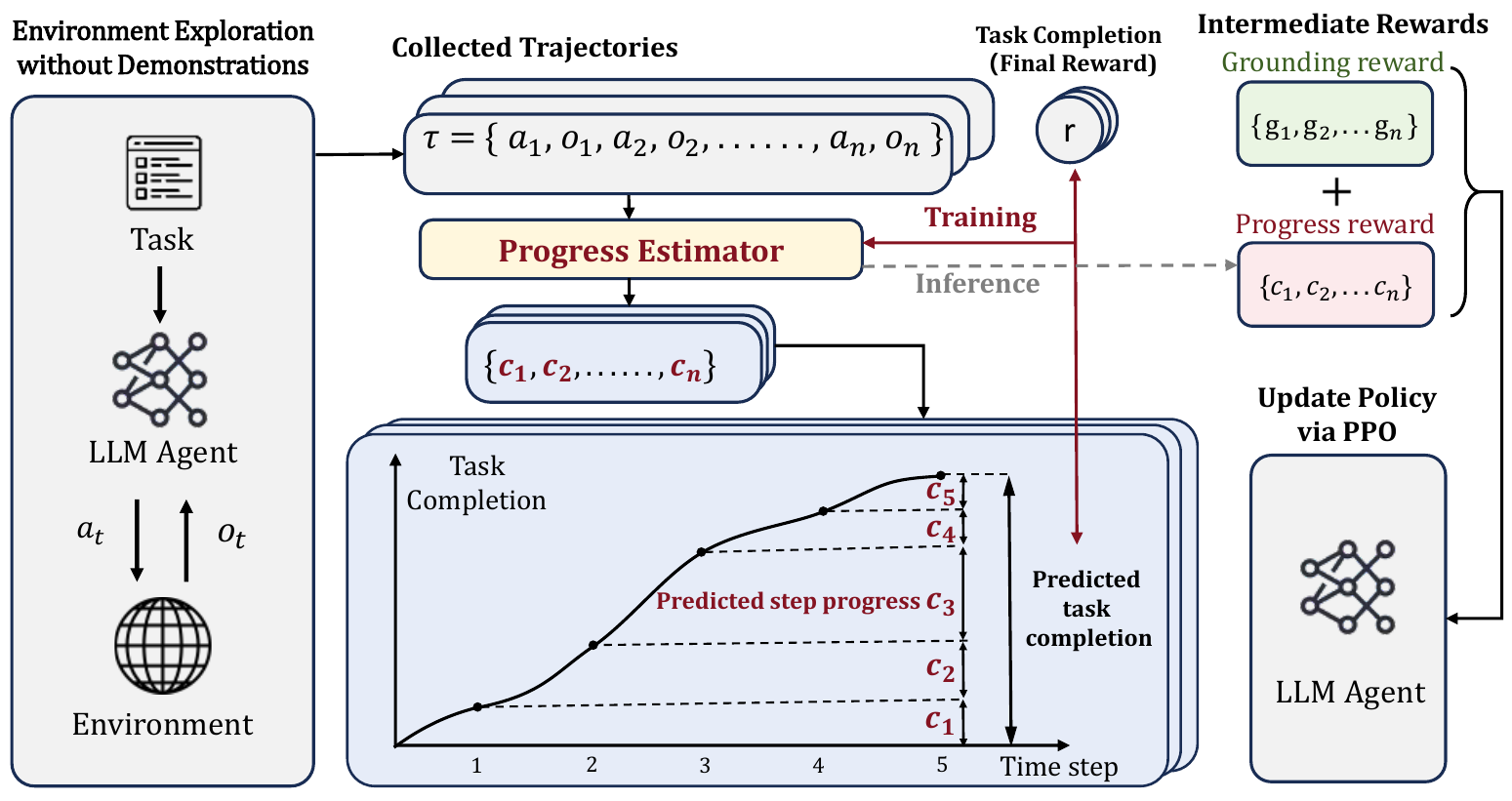}
  \vspace{-15pt}
  \caption{Overview of our proposed Stepwise Progress Attribution (SPA) framework for training LLM agents with RL.}
  % \vspace*{-1ex}
  \label{pic:framework}
\end{figure}

This section presents our proposed method, with the overview shown in Figure~\ref{pic:framework}. First, we endow LLM agents with the basic task planning capability via behavior cloning (Section~\ref{method_bc}). Then, we construct a progress estimator to attribute the final reward to stepwise intermediate rewards (Section~\ref{method_step_attribution}). Finally, we leverage these rewards for reinforcement learning of LLM agents (Section~\ref{PPO_shaped_reward}).

\subsection{Behavior Cloning}
\label{method_bc}

To achieve basic task planning, we ask LLM agents to perform behavior cloning (BC) via supervised fine‐tuning (SFT) on successful expert trajectories. Following previous studies~\citep{song2024trial,wang2025steca}, We adopt the ReAct-style~\citep{yao2023react} trajectory format, where a Chain-of-Thought (CoT) rationale~\citep{wei2022chain} is provided prior to each action. 
Since each rationale and its corresponding action are generated together, we denote the combined thought–action pair as a single unit \(a_t\) for simplicity.

Concretely, given an expert trajectory dataset $\mathcal{D}=\bigl\{(u,e)^{(i)}\bigr\}_{i=1}^{|\mathcal{D}|}$, we train the policy model by minimizing the negative log-likelihood of the expert thought-action pairs, excluding observations from the loss computation. 
After SFT, the LLM agent acquires fundamental task-planning capabilities, therefore resulting in a base agent $\pi_{base}$. This process significantly accelerates downstream policy optimization and reduces the sample complexity in the following RL training.

\subsection{Reward Redistribution via Stepwise Progress Attribution}
\label{method_step_attribution}

\paragraph{Trajectory Collection.}
Accurately estimating each action’s contribution to task completion requires observing that action across a wide range of interaction histories. To this end, we employ the base agent $\pi_{base}$ to perform large-scale exploration in the environment, thereby broadly covering the space of possible state–action sequences. 
Instead of relying on manually designed complicated exploration schemes~\citep{zhao2023large,hao2023reasoning,wang2025steca,xiong2024watch}, which constrain the action space and limit exploratory diversity, we adopt a simpler approach.
Specifically, we employ our base agent \(\pi_{\mathrm{base}}\) to conduct \(M\) rollouts for each task in the expert trajectory dataset $\mathcal{D}$ \textit{without any demonstrations}. We then aggregates the resulting trajectories to obtain the exploration dataset $\mathcal{D}_{\mathrm{explore}}=\bigl\{\bigl\{(u_i,\,\hat{\tau}_{i,j},\,R_{i,j})\bigr\}_{j=1}^{M}\bigr\}_{i=1}^{|\mathcal{D}|}$, where $\hat{\tau}_{i,j}$ represents $j$-th trajectory of $i$-th task, $R_{i,j}$ denotes its corresponding final reward.

\paragraph{Progress Estimator Training.}
Unlike previous work that evaluates step-level rewards locally, we aim to construct stepwise rewards from a global task completion perspective. Specifically, we aim to measure the contribution of per-step action to the overall task completion, which we term \textit{stepwise progress}. 
We treat the delayed reward as an indicator of the final task completion, then leverage the powerful reasoning capability of LLMs to construct a \textbf{progress estimator} $\mathcal{E}_\gamma$. It systematically decomposes the delayed reward into stepwise contributions, capturing how each action incrementally advances the agent towards task completion.
Specifically, we append a lightweight multilayer perceptron (MLP)~\citep{taud2017multilayer} to the last hidden layer of a pre-trained LLM \(\pi_{\gamma}\), enabling it to output a scalar contribution score for each state–action pair:
\begin{equation}
\hat{c}_{t}
= \mathrm{MLP}\bigl(h_t\bigr)\,, 
\quad
h_t = f_{\pi_{\gamma}}(s_t,a_t)\,,
\end{equation}
where $s_t$ represents the observed trajectory state $(u,\hat{a}_1,\hat{o}_1,...,\hat{a}_{t-1},\hat{o}_{t-1})$ at step $t$, \(f_{\pi_{\gamma}}\) denotes the encoding operation of \(\pi_{\gamma}\), and \(\hat{c}_t\) quantifies the estimated contribution of action \(a_t\) at $t$-th step toward the overall task completion.

To achieve progress estimation, we assign a scalar contribution score to each action in the trajectory and constrain that the sum of these contributions matches the ground truth task completion. Concretely, we define the predicted task completion for a trajectory $\tau$ as:
\begin{equation}
    \hat{R}
    = \sum_{t=1}^{N} \hat{c}_{t}\,,
\end{equation}
where $\hat{c}_{t}$ and $N$ denote the predicted $t$-th step contribution and the total number of steps in the trajectory $\tau$, respectively. 
To train the progress estimator $\mathcal{E}_\gamma$, we minimize the Mean Squared Error (MSE) loss between predictions and observations as follows:
\begin{equation}
\mathcal{L}_{\text{PE}}
= \frac{1}{|\mathcal{D}|*\,M}
  \sum_{i=1}^{|\mathcal{D}|}
  \sum_{j=1}^M
  \bigl(\hat{R}_{i,j} - R_{i,j}\bigr)^2,
\end{equation}
where $\hat{R}_{i,j}$ and $R_{i,j} \in [0, 1]$ denote the predicted task completion score and the observed final reward in the $j$-th rollout of the $i$-th task, respectively.
This objective encourages the model to learn to redistribute the final reward across each step in proportion to its actual contribution to task completion.

\paragraph{Stepwise Progress Prediction.}

Once the progress estimator is trained, we apply it to a given state–action pair \((s_t,a_t)\) in each trajectory \(\tau\in\mathcal{D}_{\mathrm{explore}}\), yielding per‐step contribution scores \(\{c_t\}_{t=1}^n\).  
Each \(c_t\) quantifies how much action \(a_t\) contributes to the eventual task completion. Crucially, since
$\sum_{t=1}^n c_t=\hat{R} \;\approx\; R$, the sum of per‐step contributions is expected to recover the delayed reward \(R\), ensuring that the redistributed rewards preserve the overall outcome while providing dense, fine‐grained feedback. More detailed discussions are provided in Appendix~\ref{reason_resa}.

\subsection{Reinforcement Learning with Intermediate Rewards}
\label{PPO_shaped_reward}

While obtaining the per‐step contribution score $c_t$ from the progress estimator \(\mathcal{E}_\gamma\), we note that this term only captures the contribution of a step action toward eventual task completion—it does not guarantee that the action can be successfully executed in the environment. To ensure that each intermediate reward also reflects action executability, we introduce a \textit{grounding signal} $g_t$ that takes the value 1 if $a_t$ can be successfully executed in the environment and 0 otherwise. Finally, we define the fused immediate reward as:
\begin{equation}
r_t^{\text{fused}} = \alpha\,c_t \;+\;\beta\,g_t,
\end{equation}
where \(\alpha,\beta>0\) are the hyperparameters to balance contribution versus grounding. Injecting \(r_t^{\mathrm{fused}}\) into the RL training objective yields both task‐success and environment‐grounded learning signals.

Building on the intermediate fused reward \(r_t^{\mathrm{fused}}\), we adapt the standard PPO~\citep{schulman2017proximal} algorithm to effectively leverage our dense, progress-driven, and environment-grounded feedback. Concretely, we replace the sparse terminal reward \(r_t\) by \(r_t^{\mathrm{fuse}}\) throughout the policy optimization and advantage estimation. The modified expected return is formulated as:
\begin{equation}
J_{\mathrm{fuse}}(\theta)
= \mathbb{E}_{\tau\sim\pi_\theta}\Bigl[\sum_{t=1}^n \gamma^{\,t-1}\,r_t^{\mathrm{fused}}\Bigr],
\end{equation}
and the clipped surrogate objective is unchanged in form:
\begin{equation}
\mathcal{L}_{\mathrm{fused}}^{\mathrm{CLIP}}(\theta)
= \mathbb{E}_t\Bigl[\min\bigl(\dfrac{\pi_\theta(a_t\mid s_t)}{\pi_{\theta_{\mathrm{old}}}(a_t\mid s_t)},\hat A_t^{\mathrm{fused}},\;
\mathrm{clip}(\dfrac{\pi_\theta(a_t\mid s_t)}{\pi_{\theta_{\mathrm{old}}}(a_t\mid s_t)},1-\epsilon,1+\epsilon)\,\hat A_t^{\mathrm{fused}}\bigr)\Bigr],
\end{equation}
where the advantages
\(\hat A_t^{\mathrm{fused}}\) is computed via GAE using temporal‐difference errors
\(\delta_t^{\mathrm{fused}} = r_t^{\mathrm{fused}} + \gamma\,V_\phi(s_{t+1}) - V_\phi(s_t)\), following Eq. (\ref{eq:advantage}).
By injecting \(r_t^{\mathrm{fused}}\) instead of the sparse \(r_t\), we obtain a dense reward stream that both respects stepwise progress and action executability. This integration enhances credit assignment and policy optimization, distinguishing our approach from vanilla PPO, which relies solely on sparse end‐of‐episode feedback.

%% file: sections/4_experiment.tex
\section{Experiments}
\label{sec:exp}

\begin{table}[t!]
\caption{Evaluation results on ALFWorld's unseen tasks. These results are shown for each of the six task categories as well as the overall average. ``Succ.'' denotes the task success rate,  and ``Gro.'' denotes the grounding accuracy (see Section~\ref{exp_setup}). Higher values indicate better performance.}
\vspace{5pt}
\label{tab:performance_alfworld}
\centering
\resizebox{1.0\textwidth}{!}{
\begin{tabular}{l cccccccc}
\toprule
\multirow{2}{*}{ \textbf{Method}} & \multicolumn{2}{c}{\textbf{Overall Tasks}} & \textbf{PICK} & \textbf{CLEAN} & \textbf{HEAT} & \textbf{COOL} & \textbf{LOOK} & \textbf{PICK2}\\
& Succ.(\%) & Gro.(\%) & Succ.(\%) & Succ.(\%) & Succ.(\%) & Succ.(\%) & Succ.(\%) & Succ.(\%) \\
\midrule
SFT~\citep{chen2023fireact} & 73.1 & 89.9 & 79.2 & 77.4 & 73.9 & 61.9 & 83.3 & 58.8 \\
RFT~\citep{yuan2023scaling} & 74.6 & 91.2 & 83.3 & 83.9 & 78.3 & 66.7 & 77.8 & 47.1 \\
PPO~\citep{schulman2017proximal} & 73.9 & 90.4 & 83.3 & 87.1 & 73.9 & 61.9 & 77.8 & 47.1\\
ArCHer~\citep{zhou2024archer} & 75.4 & 88.6 & 83.3 & 83.9 & 65.2 & \underline{85.7} & 66.7 & 58.8 \\
StepAgent~\citep{deng2024novice} & 75.4 & 91.5 & 83.3 & 87.1 & 78.3 & 71.4 & 77.8 & 41.2\\
RAGEN~\citep{wang2025ragen} & 75.4 & 90.1 & 91.7 & 77.4 & 78.3 & 57.1 & 88.9 & 52.9\\
PRM4A~\citep{choudhury2025process} & 73.9 & 90.8 & 58.3 & 80.6 & 73.9 & 71.4 & \underline{100.0} & 58.8\\
\midrule
\textbf{SPA (Ours)} &  \underline{79.1} & 93.7 & \underline{95.8} & 83.9 & \underline{87.0} & 61.9 & 77.8 & 58.8 \\
w/o grounding signal & 77.6 & 91.1 & 83.3 & \underline{90.3} & 65.2 & 66.7 & 88.9 & \underline{64.7} \\
w/o stepwise progress & 77.6 & \underline{94.2} & 87.5 & 80.6 & 78.3 & 66.7  & 88.9 & 58.8 \\
\bottomrule
\end{tabular}
}
% \vskip -0.1in
\end{table}

In this section, we conduct extensive experiments to validate the effectiveness of our method. The results demonstrate superior performance compared to baseline methods across various environments. Additionally, we perform ablation studies to highlight the effectiveness of each component in our intermediate reward design.

\subsection{Experimental Setup}
\label{exp_setup}

\paragraph{Benchmarks.}
We evaluate our approach on three benchmark environments: \textbf{WebShop}~\citep{yao2022webshop} for web navigation, and two embodied household suites: \textbf{ALFWorld}~\citep{shridhar2020alfworld} and \textbf{VirtualHome}~\citep{puig2018virtualhome}.
In all of these environments, the interaction between the agent and the environment can be formulated as partially observable Markov decision processes. For each task in both environments, the agent need to conduct multiple actions while interacting with the environment and at the end of the trajectory, these environments would provide a final reward to indicate whether the task is completed. For more details of our evaluated environments and corresponding datasets, please refer to Appendix~\ref{app:eval_env}.

% training setup
\paragraph{Training Setup.}

We primarily use Llama-3.2-3B-Instruct as the base model for constructing both the LLM agents and the progress estimator. During the behavior cloning stage, the policy model is trained for 3 epochs, while the progress estimator is trained for 1 epoch using all the exploration data, \(D_{\text{explore}}\).
For exploration, the number of rollouts \(M\) for the LLM agent is set to 10, with a decoding temperature of 0.7 to ensure diverse explored trajectories. The hyperparameters \(\alpha\) and \(\beta\), which balance contribution and grounding in the shaped fused reward, are set to 1 and 0.5, respectively.
In the reinforcement training stage, we employ LoRA~\citep{hu2022lora} for efficient training. For additional implementation details, please refer to Appendix~\ref{app:implem}.

% evaluation setup
% ALFWorld and Webshop/ScienceWorld; evaluation metrics
\paragraph{Evaluation Protocol.} 

When evaluating the agent, we apply ReAct-style interaction format~\citep{yao2023react} with thought generated before the action. Additionally, we add a one-shot in-context example in the instruction prompt for each task. Relevant prompt templates are provided in Appendix~\ref{app:prompt}. The decoding temperature of the LLMs is set to 0 for deterministic generation. Following previous practices~\citep{song2024trial, xiong2024watch}, we mainly adopt \textbf{success rate} as the metric to evaluate the planning capability of LLM agents. We also report the \textbf{grounding accuracy}, which is defined as the average percentage of executable actions per trajectory, to evaluate the agent's environment grounding capability.

\paragraph{Baselines.}

% version1
% 我们比较了
We evaluate our method against the following baseline methods: 
(1) \textbf{SFT}~\citep{chen2023fireact,zeng2023agenttuning} conducts behavior cloning on expert trajectories.
(2) \textbf{RFT}~\citep{yuan2023scaling} adds the success trajectories from rejection sampling to the expert trajectories and trains the agent on the augmented data with fine-tuning.
(3) \textbf{PPO}~\citep{schulman2017proximal} is an RL method directly optimizing the SFT agents to maximize the final reward.
(4) \textbf{ArCHer}~\citep{zhou2024archer} is a hierarchical multi-turn RL framework designed for training LLM agents.
(5) \textbf{StepAgent}~\citep{deng2024novice} optimizes agent policy by incorporating stepwise supervision into reinforcement learning, thereby approaching the expert policy.
(6) \textbf{RAGEN}~\citep{wang2025ragen} introduces a bi-level GAE approach, which enables more fine-grained rewards for RL training between different interaction turns.
(7) \textbf{PRM4A}~\citep{choudhury2025process} utilizes rollout trajectories from exploration to train a process reward model, which offers a process-level signal for policy training.

\subsection{Overall Results}
\label{sec:overall_results}

\begin{figure}[t]
    \centering
    % 第一张子图
    \begin{subfigure}{0.32\textwidth} % 正确的宽度设置
        \centering
        \includegraphics[width=\textwidth]{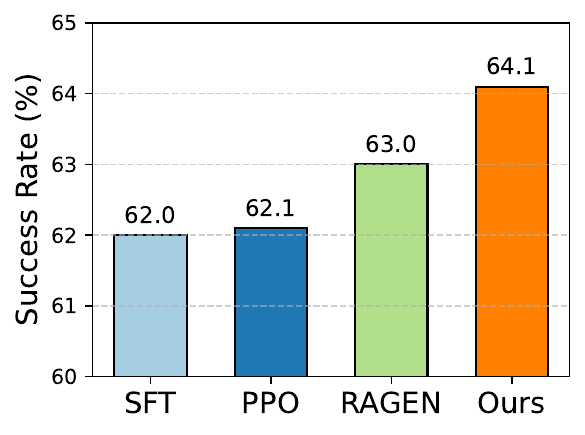}
        \caption{Success rate in the WS.}
        \label{fig:success_rate}
    \end{subfigure}
    \hfill
    % 第二张子图
    \begin{subfigure}{0.32\textwidth} % 正确的宽度设置
        \centering
        \includegraphics[width=\textwidth]{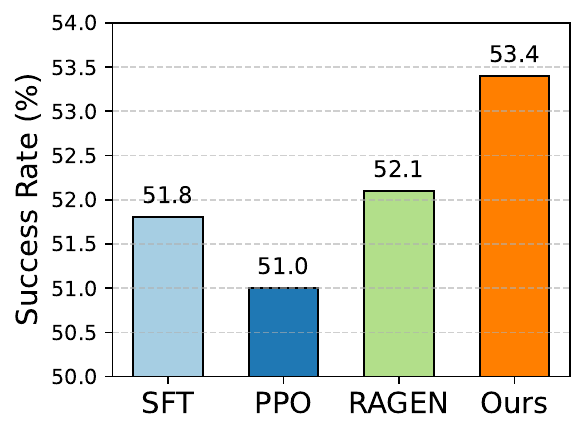}
        \caption{Success rate in the VH.}
        \label{fig:grounding_accuracy}
    \end{subfigure}
    \hfill
    % 第三张子图
    \begin{subfigure}{0.32\textwidth} % 正确的宽度设置
        \centering
        \includegraphics[width=\textwidth]{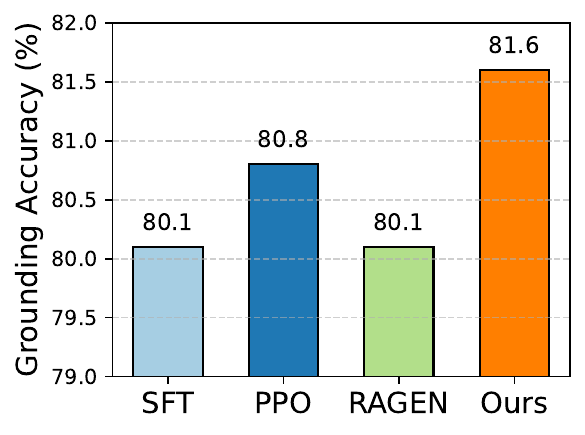}
        \caption{Grounding accuracy in the VH.}
        \label{fig:comparison_virtualhome}
    \end{subfigure}
    % 总图标题
    \caption{Performance comparison of various methods across different evaluation metrics and environments. ``WS'' and ``VH'' denote the Webshop and VirtualHome environments, respectively. 
    In the WebShop environment, where the action set is predefined for each step and provided by the environment, the grounding accuracy consistently reaches 100\%, owing to the powerful instruction-following capabilities of LLMs, and thus we do not report it here.
    }
    \label{fig:combined_performance}
\end{figure}

% version2.2
In this section, we present the overall results of our method compared to baseline approaches across various environments and evaluation metrics.
As shown in Table~\ref{tab:performance_alfworld}, our method achieves the highest overall performance in unseen ALFWorld tasks, with a task success rate of 79.1\%, which is 3.6\% higher than the state-of-the-art method, StepAgent. Additionally, our approach reaches a grounding accuracy of 91.7\%, surpassing all baseline methods. Notably, it outperforms across five out of six task categories, with significant improvements in the PICK and CLEAN categories, where success rates increased to 95.8\% and 90.3\%, respectively. These results underscore the effectiveness of our reward redistribution strategy in enhancing both task success and grounding accuracy.
Ablation studies further confirm the effectiveness of our reward components. The contribution reward improves performance by accurately capturing step-wise contributions, while the grounding reward enhances execution accuracy through robust guidance. When combined, these rewards work synergistically to significantly boost overall performance, highlighting the importance of their integration.
Beyond the ALFWorld environment, we evaluated our method in WebShop and VirtualHome, as illustrated in Figure~\ref{fig:combined_performance}. In WebShop, our approach achieved a success rate of 64.1\%, outperforming the closest baseline, RAGEN, by 1.1\%. In the more challenging VirtualHome environment, characterized by tasks with longer horizons, our method achieved a success rate of 53.4\%, surpassing RAGEN by 1.3\%, while also attaining the highest grounding accuracy of 81.6\%. These results demonstrate the robustness of our method across diverse environments and task complexities, driven by effective reward decomposition.

%% file: sections/5_analyses.tex
\section{Analyses and Discussions}
\label{sec:analyses}

\subsection{Effectiveness Validation of the Progress Estimator}

% version2
To verify that our trained progress estimator generates meaningful intermediate rewards rather than random noise, we evaluate five reward variants based on PPO in the ALFWorld environment. In our proposed SPA, the reward at each time step is derived from the stepwise progress reward and grounding reward. To assess the effectiveness of our method, we compare it against several alternative reward mechanisms.
In the “MC” condition, the learned signals are replaced with a Monte Carlo rollout estimate of the current step reward. The “Random” condition introduces uniform random values within the range $[0,1]$ as in-time rewards, effectively injecting noise into the system. The “Mean” condition distributes the terminal reward equally across all steps, ignoring step-specific contributions. Lastly, the standard PPO baseline relies solely on the delayed terminal reward without incorporating any intermediate rewards. 

\begin{figure}[t]
    \centering
    % 第一幅图
    \begin{minipage}{0.49\textwidth}
        \centering
        \includegraphics[width=\textwidth]{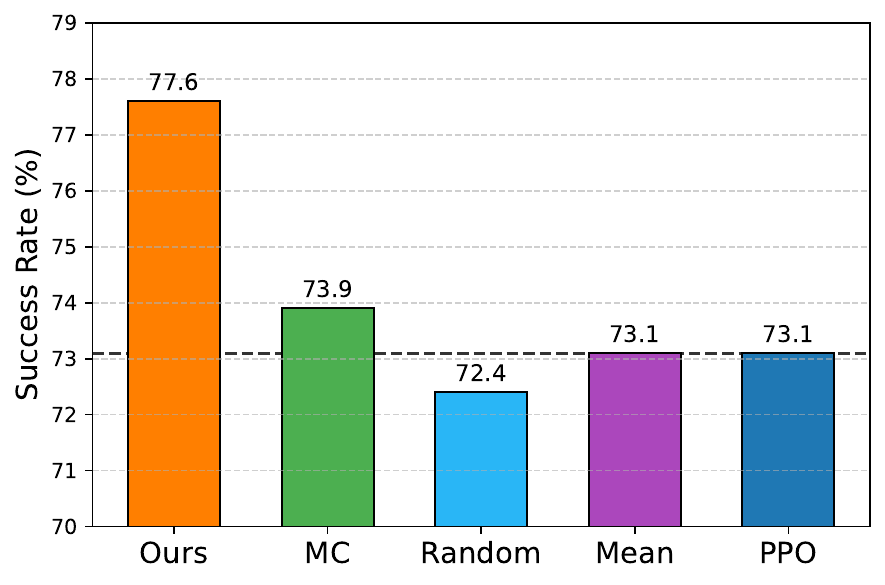}
        \caption{Performance with different intermediate rewards in the ALFWorld environment.}
        \label{fig:vali_estimator}
    \end{minipage}
    \hfill
    % 第二幅图
    \begin{minipage}{0.49\textwidth}
        \centering
        \includegraphics[width=\textwidth]{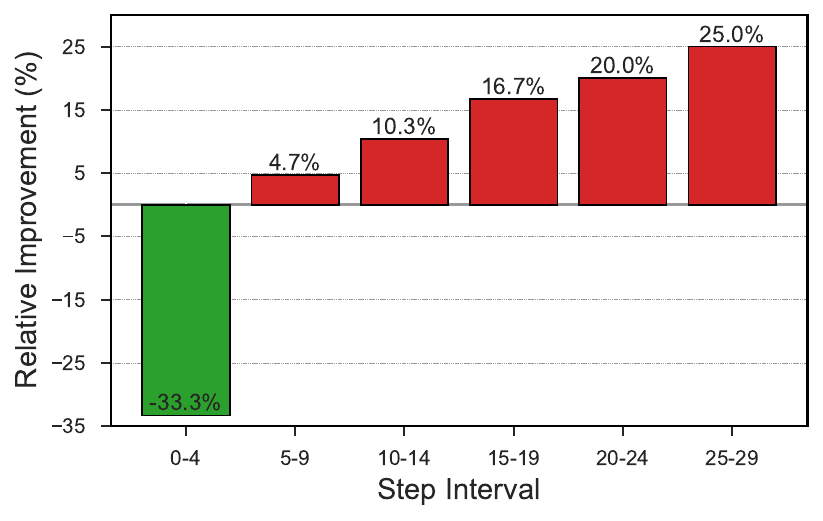}
        \caption{Relative performance improvement between our SPA and PPO at different step intervals.}
        \label{fig:multi_step_eval}
    \end{minipage}
\end{figure}

Figure~\ref{fig:vali_estimator} illustrates the success rates of the five reward variants. Our method outperforms all alternatives, achieving the highest success rate and surpassing the strongest heuristic, MC, by a notable margin. In contrast, both the Random and Mean conditions perform similarly to or worse than the standard PPO baseline, showing that arbitrary or uniformly distributed intermediate rewards fail to improve performance and can even be detrimental.
These results highlight the effectiveness of the progress estimator in providing meaningful, well-calibrated intermediate rewards. By offering denser and more informative reward signals, our approach significantly improves credit assignment in long-horizon planning tasks, whereas naïve alternatives contribute little to no benefit.

\subsection{Analyses of Long-horizon Task Completion}

% version2
To evaluate the impact of SPA on long-horizon tasks, we compare its performance with PPO across varying step intervals in the ALFWorld test set. 
Specifically, the number of steps required by the LLM agent to complete different tasks varies, and we group tasks into different step intervals, counting the number of tasks completed in each interval and calculating the relative improvement of SPA over PPO in terms of task completion.
As shown in Figure~\ref{fig:multi_step_eval}, SPA consistently outperforms PPO as the step interval increases, with a notable 25\% improvement in the 25 $\sim$ 29 step range. This trend highlights SPA’s enhanced ability to handle long-horizon scenarios by effectively attributing rewards to intermediate actions, thereby improving credit assignment over extended trajectories.
Interestingly, SPA exhibits a relative performance drop in the 0 $\sim$ 4 step interval, likely due to the simplicity of short-horizon tasks where exploration is unnecessary, and PPO's reliance on terminal rewards suffices. This suggests that SPA's strength lies in its ability to support complex, multi-step planning, particularly when long-term exploration and exploitation are critical.

\subsection{Discussions on Different Methods in Terms of Credit Assignment}

\begin{table}[t]
\caption{Comparison of different LLM agent training methods based on credit assignment types, granularities, enhancement strategies, and their success rates in the ALFWorld environment. 
}
\vspace{5pt}
\label{tab:credit_assignment_sorted}
\centering
\resizebox{0.94\textwidth}{!}{
\begin{tabular}{l ccccc}
\toprule
\textbf{Method} & \textbf{Type} &  \textbf{Granularity}  &  \textbf{Enhancement Strategy}   & \textbf{Success (\%)} \\
\midrule
\textbf{SPA (Ours)} & GAE & Token & Reward Redistribution & \textbf{79.1} \\
\midrule
RAGEN~\citep{wang2025ragen} & GAE & Token & Hierarchical Decomposition & 75.4 \\
PRM4A~\citep{choudhury2025process} & GAE & Token & Reward Shaping & 73.9 \\
StepAgent~\citep{deng2024novice}  & GAE & Token & Reward Shaping & 75.4 \\
PPO~\citep{schulman2017proximal} & GAE & Token &  None &  73.9 \\
% \hdashline
RLOO~\citep{ahmadian2024back} & RLOO & Trajectory & None &  71.6 \\
GRPO~\citep{guo2025deepseek} & GRPO & Trajectory & None &  70.9 \\
% \hdashline
Policy Gradient~\citep{sutton1999policy} & None & Trajectory & None & 72.4 \\
\bottomrule
\end{tabular}
}
\end{table}

% Credit assignment is critical in reinforcement learning, particularly in tasks with delayed rewards. 
Addressing delayed rewards in agentic tasks is closely related to a typical problem in RL: \textit{credit assignment}. It refers to the problem of determining which actions or decisions in a sequence were responsible for a specific outcome, especially when the reward is received much later than the action that caused it.
Table~\ref{tab:credit_assignment_sorted} summarizes different methods based on their credit assignment types, granularities, and enhancement strategies.
We conclude that \textbf{granularity} plays a key role in performance. Trajectory-level methods, such as GRPO and RLOO, assign credits to an entire trajectory but perform worse than Policy Gradient, which lacks explicit credit assignment. This highlights the limitations of coarse-grained credit assignment. In contrast, token-level methods, including RAGEN, PRM4A, and ours, consistently achieve higher success rates, showcasing the necessity of fine-grained credit assignment. However, token-level approaches face optimization inefficiencies due to their high complexity. This suggests that step-level credit assignment, which balances granularity and efficiency, is a promising direction.
\textbf{Enhancement strategies} also impact training process. Methods like RAGEN and PRM4A introduce intermediate rewards or hierarchical decomposition to address delayed rewards, which is effective. Our method further ensures consistency between intermediate and delayed rewards through reward redistribution, achieving the highest success rate. However, relying on additional models for intermediate rewards introduces potential bias. Future research could explore model-free reward redistribution and step-level granularity to improve scalability and robustness.

%% file: sections/6_related_work.tex
\section{Related Work}

\paragraph{LLM Agent Learning.}

% version2
To address agentic tasks, existing LLM agent learning approaches can be broadly divided into two categories: supervised fine-tuning methods and reinforcement learning (RL)-based methods.
Supervised fine-tuning typically relies on collecting large-scale expert trajectories to enhance the planning capabilities of agents~\citep{chen2024agent, chen2023fireact, zeng2023agenttuning}. Additionally, some studies designed manual exploration strategies to construct reflection trajectories, enabling agents to develop self-reflection and better adapt to robust and dynamic environments~\citep{wang2024e2cl, yuan2025agent}.
Reinforcement learning (RL)-based methods, such as Proximal Policy Optimization~\citep{schulman2017proximal}, enable LLM agents to learn directly from interaction experiences~\citep{song2024trial, xiong2024watch, deng2024novice}. Unlike supervised fine-tuning, RL naturally aligns with the objective of maximizing cumulative rewards through agent-environment interactions. This makes RL particularly well-suited for agentic tasks. Therefore, in this work, we focus on RL-based approaches to further enhance the capabilities of LLM agents. The key lies in learning through interactions and optimizing for long-term rewards.

\paragraph{Process Supervision in RL.}

LLM agents must interact with the environment across multiple steps before receiving a final reward at the end of the trajectory, making it challenging to assess the quality of intermediate actions~\citep{xiong2024watch, chen2025atlas}. This lack of intermediate feedback hinders RL training and often requires extensive interactions with the environment to evaluate the trajectory. 
Incorporating process supervision or intermediate signals has been verified as effective in greatly enhancing the effectiveness and efficiency of LLM agent training.
Existing methods have extensively explored rule-based approaches~\citep{yu2024steptool,dou2024stepcoder} to shape intermediate rewards, Monte Carlo sampling to estimate the value of the current step~\citep{choudhury2025process, setlur2024rewarding,qu2025optimizing,xiong2024watch}, and using imitation of expert policies as rewards~\citep{deng2024novice}. However, these methods primarily focus on optimizing local actions and lack a holistic perspective on task completion. 
To address this, we propose measuring the progress of stepwise actions toward overall task completion as intermediate rewards, offering a more goal-oriented signal for RL.

%% file: sections/7_conclusion.tex
\section{Conclusion}

In this work, we propose Stepwise Progress Attribution (SPA), a novel framework designed to tackle the challenge of sparse and delayed rewards in reinforcement learning. 
SPA redistributes the final reward into fine-grained, stepwise signals, enabling more precise credit assignment by aligning each incremental action with overall task progress.
We validate the effectiveness of SPA through extensive experiments on representative agent benchmarks. Across all tasks, SPA consistently achieves state-of-the-art performance, significantly improving both success rates and grounding accuracy.
These results demonstrate that by offering more informative intermediate feedback, SPA provides a practical and scalable solution for training LLM agents on complex, long-horizon tasks.

%% file: sections/appendix.tex
\section{Theoretical Analyses of the Progress Estimation}
\label{reason_resa}

% version5
In this section, we analyze the proposed reward decomposition approach, where the \textbf{progress estimator} \( \mathcal{E}_\gamma \) is trained to predict stepwise contributions. We demonstrate that this approach preserves the original policy optimization objective and does not alter the policy gradient, ensuring theoretical consistency.

% ### Reward Decomposition Using Progress Estimator

Consider an episodic MDP (or POMDP over interaction histories) with a state (or history) sequence \( e_0 \to e_1 \to \cdots \to e_n \), actions \( a_t \in \mathcal{A} \), transition dynamics \( P(e_t \mid e_{t-1}, a_t) \), and a sparse reward function defined as:
\begin{equation}
r_t =
\begin{cases}
0, & t < n, \\
r \in [0, 1], & t = n.
\end{cases}
\end{equation}

To replace the sparse terminal reward \( r \) with dense, stepwise feedback, we introduce a **Progress Estimator** \( \mathcal{E}_\gamma \). The estimator predicts the per-step contribution \( \hat{c}(s_t, a_t) \), which quantifies the contribution of each action \( a_t \) in state \( s_t \) to the overall task completion. Specifically, we define:
\begin{equation} \label{eq:progress_estimator}
\hat{c}(s_t, a_t) = \mathrm{MLP}(\mathbf{h}_t), \quad \mathbf{h}_t = f_{\pi_\gamma}(s_t, a_t),
\end{equation}
where \( f_{\pi_\gamma} \) extracts the contextual embedding of the state-action pair \( (s_t, a_t) \) from the pre-trained policy model \( \pi_\gamma \). 

The predicted contribution \( \hat{c}(s_t, a_t) \) decomposes the reward into dense, stepwise signals using a potential function \( \Phi_\psi \), such that:
\begin{equation} \label{eq:constructed_reward}
\hat{c}(s_t, a_t) = \Phi_\psi(e_t) - \Phi_\psi(e_{t-1}),
\end{equation}
where \( \Phi_\psi \) assigns a scalar potential value to each interaction history \( e_t \). By directly predicting the difference \( \Phi_\psi(e_t) - \Phi_\psi(e_{t-1}) \) using \( \mathcal{E}_\gamma \), we bypass the need to explicitly compute \( \Phi_\psi(e_t) \), making the training efficient and focused on the stepwise contributions.

The cumulative reward under this decomposition is:
\begin{equation} \label{eq:cumulative_reward}
\hat{R} = \sum_{t=1}^n \hat{c}(s_t, a_t) = \Phi_\psi(e_n) - \Phi_\psi(e_0).
\end{equation}
By enforcing \( \Phi_\psi(e_0) = 0 \), we ensure \( \hat{R} = \Phi_\psi(e_n) \). To align \( \hat{R} \) with the ground truth reward \( r \), we train \( \mathcal{E}_\gamma \) to minimize the following objective.

% ### Training the Progress Estimator

The Progress Estimator \( \mathcal{E}_\gamma \) is trained to predict contributions \( \hat{c}(s_t, a_t) \) such that the cumulative reward \( \hat{R} \) matches the observed ground truth reward \( r \). This is achieved by minimizing the **sum-of-deltas loss** over trajectories \( \tau \sim D_{\text{expl}} \):
\begin{equation} \label{eq:loss_sum_deltas}
\mathcal{L}_{\text{tel}}(\gamma) = \mathbb{E}_{\tau \sim D_{\text{expl}}} \bigl[\bigl(\hat{R} - r\bigr)^2\bigr],
\end{equation}
where \( \hat{R} = \sum_{t=1}^n \hat{c}(s_t, a_t) \). Under standard assumptions (e.g., sufficient model capacity, i.i.d.\ sampling, and convexity of the expected risk), the minimizer of \( \mathcal{L}_{\text{tel}} \) satisfies:
\begin{equation}
\hat{R} = \sum_{t=1}^n \hat{c}(s_t, a_t) \to \mathbb{E}[r \mid e_n].
\end{equation}
This ensures that the cumulative contributions predicted by \( \mathcal{E}_\gamma \) align with the observed task completion score \( r \).

% ### Policy Gradient Equivalence

Using the dense, stepwise contributions \( \hat{c}(s_t, a_t) \) in place of the sparse terminal reward \( r \), we compute the policy gradient under the REINFORCE objective~\citep{sutton1999policy}:
\begin{equation}
\nabla_\theta J(\theta) = \mathbb{E} \Bigl[\sum_{t=1}^n \nabla_\theta \log \pi_\theta(a_t \mid e_{t-1}) G_t \Bigr],
\end{equation}
where \( G_t = \sum_{k=t}^n r_k \) is the return from step \( t \) onward. Substituting the constructed reward \( \hat{c}(s_t, a_t) \), the return becomes:
\begin{equation}
G_t = \sum_{k=t}^n \hat{c}(s_k, a_k) = \Phi_\psi(e_n) - \Phi_\psi(e_{t-1}).
\end{equation}

The policy gradient then becomes:
\begin{equation}
\nabla_\theta J(\theta) = \mathbb{E} \Bigl[\sum_{t=1}^n \nabla_\theta \log \pi_\theta(a_t \mid e_{t-1}) \bigl(\Phi_\psi(e_n) - \Phi_\psi(e_{t-1})\bigr)\Bigr].
\end{equation}
Since \( \Phi_\psi(e_n) \) depends only on the terminal state \( e_n \), terms involving \( \Phi_\psi(e_{t-1}) \) telescope to zero in expectation. Therefore, the gradient simplifies to:
\begin{equation}
\nabla_\theta J(\theta) = \mathbb{E} \Bigl[\sum_{t=1}^n \nabla_\theta \log \pi_\theta(a_t \mid e_{t-1}) G_t \Bigr],
\end{equation}
which is identical to the gradient under the original sparse reward. This demonstrates that training \( \mathcal{E}_\gamma \) and using the predicted contributions \( \hat{c}(s_t, a_t) \) does not alter the policy optimization objective.

% ### Summary

In summary, the progress estimator \( \mathcal{E}_\gamma \) directly predicts stepwise contributions \( \hat{c}(s_t, a_t) \), which decompose the terminal reward \( r \) into dense, meaningful feedback. By predicting \( \hat{c}(s_t, a_t) = \Phi_\psi(e_t) - \Phi_\psi(e_{t-1}) \), the estimator avoids explicitly computing the potential values \( \Phi_\psi \) while ensuring that the cumulative reward \( \hat{R} \) matches \( r \). Crucially, this reward decomposition does not alter the policy optimization objective, preserving the equivalence of the policy gradient updates while enabling dense credit assignment.

\section{PPO's Limitations in Addressing Delayed Rewards}
\label{more_theo_proof}

Assume that the value network \(V_\phi\) has converged to the true discounted return
\begin{equation}
  V_\phi(s_t) = \mathbb{E}\!\Bigl[\sum_{k=t+1}^n \gamma^{\,k-t-1}\,r_k\Bigr].
\end{equation}
In the sparse‐reward setting, where \(r_k=0\) for all \(k<n\) and only \(r_n\in\{0,1\}\) may be nonzero, this expectation collapses to
\begin{equation}
  V_\phi(s_t) = \gamma^{\,n-t-1}\,\mathbb{E}[r_n], 
  \quad 
  V_\phi(s_n) = 0.
\end{equation}

By definition, the one‐step TD‐error at time \(t\) is
\begin{equation}
  \delta_t = r_{t+1} + \gamma\,V_\phi(s_{t+1}) - V_\phi(s_t).
\end{equation}
For each intermediate step \(t=1,\dots,n-2\), since \(r_{t+1}=0\) and
\begin{equation}
  \gamma\,V_\phi(s_{t+1})
  = \gamma\bigl[\gamma^{\,n-(t+1)-1}\,\mathbb{E}[r_n]\bigr]
  = \gamma^{\,n-t-1}\,\mathbb{E}[r_n]
  = V_\phi(s_t),
\end{equation}
it follows that
\begin{equation}
  \delta_t = 0,\quad t = 1,\dots,n-2.
\end{equation}
At the penultimate step \(t=n-1\), using \(V_\phi(s_n)=0\) one obtains
\begin{equation}
  \delta_{n-1}
  = r_n + \gamma\,V_\phi(s_n) - V_\phi(s_{n-1})
  = r_n - \mathbb{E}[r_n],
\end{equation}
which in general does not vanish.

Under Generalized Advantage Estimation (GAE)~\citep{schulman2015high}, the advantage estimate becomes
\begin{equation}
  \hat A_t
  = \sum_{k=0}^{\,n-1-t} (\gamma\lambda)^k\,\delta_{t+k}
  = (\gamma\lambda)^{\,n-1-t}\,\delta_{n-1},
\end{equation}
and the resulting PPO policy gradient is
\begin{equation}
  \nabla_\phi J
  \approx
  \mathbb{E}\Bigl[\sum_{t=1}^{n-1} \nabla_\phi\log\pi_\phi(a_t\mid s_t)\,\hat A_t\Bigr]
  =
  \mathbb{E}\Bigl[\sum_{t=1}^{n-1} (\gamma\lambda)^{\,n-1-t}\,\delta_{n-1}
  \;\nabla_\phi\log\pi_\phi(a_t\mid s_t)\Bigr].
\end{equation}
Whenever \(\gamma\lambda<1\), the factor \((\gamma\lambda)^{\,n-1-t}\) decays exponentially in the temporal gap \(n-1-t\). Early actions thus receive vanishing weight in the gradient, and a single terminal reward cannot effectively propagate back to decisions made many steps before the end. Consequently, PPO with GAE proves ineffective for long‐horizon, sparse‐reward, multi‐turn tasks.

\section{Additional Details of Agent Benchmarks}
\label{app:eval_env}

\textbf{WebShop}\footnote{\url{https://github.com/princeton-nlp/WebShop}}~\citep{yao2023react} is a simulated online shopping platform designed as an environment for agents to navigate and make purchases in response to user-provided instructions. Within this environment, agents interact with the platform to evaluate and select products, ultimately executing a "buy" action when a purchase decision is made. The environment assigns a final reward upon completing the purchase, which is determined based on a set of matching heuristics that evaluate the alignment of the selected product's attributes and price with the given instructions. This setup provides a controlled setting for studying decision-making and optimization in goal-oriented agent behavior.

\textbf{ALFWorld}\footnote{\url{https://github.com/alfworld/alfworld}}~\citep{shridhar2020alfworld} offers interactive TextWorld environments that are meticulously aligned with the embodied environments introduced in ALFRED~\citep{shridhar2020alfworld}. This framework challenges agents to navigate complex household settings and execute high-level instructions, thereby testing their ability to perform practical tasks. The dataset is structured into two distinct evaluation sets: a seen set, designed to assess in-distribution generalization, and an unseen set, which comprises novel task instances to evaluate out-of-distribution generalization capabilities. At the conclusion of each trajectory, the environment provides a binary reward, indicating whether the agent has successfully completed the assigned task. This setup facilitates a clear and measurable assessment of agent performance in both familiar and novel scenarios.

\textbf{VirtualHome}\footnote{\url{https://github.com/xavierpuigf/virtualhome}}~\citep{puig2018virtualhome} is a comprehensive dataset that encompasses 292 high-level household tasks and 1,374 unique action plans across 6,201 diverse virtual environments. The dataset was meticulously curated through detailed manual annotations provided by Amazon Mechanical Turk workers, who thoroughly labeled both the tasks and their corresponding action plans. Each entry in the dataset comprises three key components: a high-level task, a descriptive explanation, and executable action programs specifically designed to be compatible with the VirtualHome environment.
To evaluate task completion, all tasks were executed within the environment, and the final state of the environment was recorded upon task completion. A task was deemed successfully completed if the state of the environment, after exploration by the LLM agent, matched the predefined target state. To ensure the dataset's high quality, only trajectories that achieved successful final outcome rewards were retained. Additionally, it was verified that every action in the planning sequence could be executed within the environment. Following prior work ~\citep{wang2025steca}, we utilized their high-quality dataset for our experiments and analyses.

\begin{table}[t]
\caption{Statistics of the benchmarks used in our experiments.}
\centering
\setlength{\tabcolsep}{6pt} % Adjust column spacing
\renewcommand{\arraystretch}{1.2} % Adjust row spacing
\begin{tabular}{lcccc}
\toprule
\textbf{Benchmark} & \textbf{Train} & \textbf{Test} & \textbf{\#Actions} & \textbf{\#Avg./Max. Turns} \\
\midrule
Webshop      & 1938   & 200  &  -    & 3.6 / 9  \\
ALFWorld     & 2,851  & 134  & 13   & 8.0 / 20 \\
VirtualHome  & 4,920  & 247  & 40   & 11.5 / 20 \\
\bottomrule
\end{tabular}
\label{tab:dataset}
\end{table}

\section{Implementation Details}
\label{app:implem}

During the construction of the base agent, we trained the model for 3 epochs. We use the AdamW optimizer~\citep{loshchilov2017decoupled} and cosine annealing learning rate scheduler to optimize the model parameters. For the warm-up stage, the batch size is 16 and the learning rate is set to 2e-5.
For the training of the progress estimator, we utilized all exploration data and trained the model for 1 epoch. The batch size was set to 8, and we used the Adam optimizer with a learning rate of 3e-6, along with the WarmupLR scheduler.
In the reinforcement learning phase, we also trained the model for only 1 epoch, setting the learning rate to 1e-5. To enable more efficient training, we adopted LoRA (Low-Rank Adaptation) for parameter-efficient fine-tuning~\citep{hu2022lora}. Specifically, only the LoRA parameters were updated during training, while the base model parameters remained frozen.
During the inference phase, we evaluated the agent's performance using greedy decoding with the temperature set to 0. To accelerate inference, we leveraged the vLLM library~\citep{kwon2023efficient} to optimize the generation process of the LLMs.

All experiments were conducted on a computational cluster equipped with 8 NVIDIA A6000 GPUs, each with 48GB of memory. Throughout our experiments, we used an open-source model, LLaMA-3.2-3B-Instruct, and strictly adhered to the licensing terms for academic use associated with this model. The licensing terms can be found at the following link: \url{https://huggingface.co/meta-llama/Llama-3.2-1B/blob/main/LICENSE.txt}.

\section{Limitations}
\label{app_limi}

% 没有进行迭代或者说是online
% 没有采用更大的模型进行scaling测试

While our approach demonstrates superior performance compared to baseline methods, it is important to acknowledge the following limitations of our current work:

(1) Dependency on Task-Specific Tuning:
Despite the strong performance of SPA across various agentic tasks, the progress estimator in our framework requires domain-specific tuning. This reliance on task-specific customization could limit the generalizability of SPA to unseen or highly diverse agentic environments, where task characteristics may differ significantly from those in our evaluated benchmarks.

(2) Scalability to Real-World Complexities:
While SPA is designed to handle long-horizon tasks effectively, its performance in real-world environments with extremely large action spaces or trajectories spanning thousands of steps remains unexplored. In such scenarios, the accuracy of the progress estimator may degrade, which could negatively impact the overall performance of the framework.

\section{Prompt Templates}
\label{app:prompt}

We provide the inference prompt for each task, which includes a general instruction, a one-shot example, the specific task instruction, and history trajectory.

\begin{tcolorbox}[title=Inference Prompt]
\textbf{\textit{\# General Instruction:}} \\
\textbf{Human}: Interact with a household to solve a task. Imagine you are an intelligent agent in a household environment and your target is to perform actions to complete the task goal...\\
Your response should use the following format: \\
Thought: <your thoughts> \\
Action: <your next action> \\
% \textbf{Agent}: OK\\

\textbf{\textit{\# In-Context Example:}} \\
\textbf{Human}: The task is Drink (Drink water). \\
... \\

\textbf{\textit{\# Task Instruction:}} \\
\textbf{Human}: The task is xxx. \\
\textit{\textbf{(History trajectory)}} \\
... \\

\end{tcolorbox}
\begin{figure}[ht]
    \centering
    % \vspace{-8pt}
    \caption{Inference prompt template for ALFWorld and VirtualHome environment.}
    \label{fig:Inference prompt2}
\end{figure}

\begin{tcolorbox}[title=Inference Prompt]
\textbf{\textit{\# General Instruction:}} \\
\textbf{Human}: You are web shopping.
I will give you instructions about what to do.
You have to follow the instructions.
Every round I will give you an observation and a list of available actions, you have to respond an action based on the state and instruction.
You can use search action if search is available.
You can click one of the buttons in clickables.
An action should be of the following structure:
search[keywords]
click[value]
If the action is not valid, perform nothing.
Keywords in search are up to you, but the value in click must be a value in the list of available actions.
Remember that your keywords in search should be carefully designed.
Thought: I think ... \\
Action: click[something] \\
% \textbf{Agent}: OK\\

\textbf{\textit{\# In-Context Example:}} \\
\textbf{Human}: i need a long lasting 6.76 fl oz bottle of l'eau d'issey, and price lower than 100.00 dollars [SEP] Search. \\
... \\

\textbf{\textit{\# Task Instruction:}} \\
\textbf{Human}: i neeed xxx. \\
\textit{\textbf{(History trajectory)}} \\
... \\

\end{tcolorbox}
\begin{figure}[ht]
    \centering
    % \vspace{-8pt}
    \caption{Inference prompt template for WebShop environment.}
    \label{fig:Inference prompt}
\end{figure}